\documentclass[conference]{IEEEtran}
\IEEEoverridecommandlockouts
\usepackage{cite}
\usepackage{amsmath,amssymb,amsfonts}
\usepackage{graphicx}
\usepackage{booktabs}
\usepackage{textcomp}
\usepackage{xcolor}
\usepackage{url}
\def\BibTeX{{\rm B\kern-.05em{\sc i\kern-.025em b}\kern-.08em T\kern-.1667em\lower.7ex\hbox{E}\kern-.125emX}}

\begin{document}

\title{HaorFloodAlert: A 72-Hour Machine Learning Early Warning System for Flash Floods in Bangladesh's Haor Wetlands}

\author{
\IEEEauthorblockN{Salma Hoque Talukdar Koli\textsuperscript{1}, Fahima Haque Talukder Jely\textsuperscript{2}, Md.\ Samiul Alim\textsuperscript{1}, and Md.\ Zakir Hossen\textsuperscript{3}}
\IEEEauthorblockA{\textsuperscript{1}Department of Computer Science and Engineering, RTM Al-Kabir Technical University, Sylhet-3100, Bangladesh\\
\textsuperscript{2}Department of Computer Science and Engineering, North East University Bangladesh, Sylhet, Bangladesh\\
\textsuperscript{3}Department of Computer Science and Engineering, Dhaka University of Engineering \& Technology, Gazipur, Bangladesh\\
Email: info.salmahoquetalukdarkoli@gmail.com}
}

\maketitle

\begin{abstract}
Every spring, flash floods strike the haor wetlands of northeast Bangladesh just before the boro rice harvest, and one flood can erase a family's entire crop in days. Warning people in time is hard here for a structural reason: the Sunamganj Haor is a flat, bowl-shaped basin that fills at once from local rain, domestic rivers, and the Barak River in India, while fewer than twelve working gauges cover its 8{,}000 km$^2$. Existing models add a quieter problem of their own, because they train on raw temperature, which simply follows the season, so they learn the calendar instead of the flood, and none of them delivers a warning to a farmer. HaorFloodAlert answers both problems with free data alone: Sentinel-1 radar that sees through storm clouds, rainfall records and forecasts, soil moisture, and a modeled upstream Barak signal worth about 36 hours of lead time. A monthly climatological anomaly then removes the seasonal bias, cutting the temperature-label correlation from $r=0.570$ to $r=-0.031$. Tested by leave-one-out cross-validation on 77 events with real Sentinel-1 images (2014--2024), the Random Forest and XGBoost ensemble reaches 90.9\% accuracy, 89.2\% F1-score, and AUC 0.939, and these labels hold up against 12.3 years of official gauge records. The same system then ran live for ten days in May--June 2026 and raised a high-risk alert about three days before the river neared its danger level. Warnings go out in Bengali by SMS, e-mail, and WhatsApp, and every number here can be regenerated from our public, seeded pipeline.
\end{abstract}


\section{Introduction}
A flash flood is a rapid rise of water within hours to a few days of its trigger, leaving little time to react; in the haor wetlands it takes a particular form, backwater flooding, where water pools in a closed basin from several directions at once instead of moving down a single channel. The haor basins of northeast Bangladesh are shallow depressions of this kind, covering roughly 8{,}000~km$^2$. They flood fully during the monsoon (June--September), drain by November, and support a single boro rice crop. That crop ripens in March--April, exactly when pre-monsoon flash floods can arrive with little warning \cite{kamal2018}. A flood two weeks before harvest triggers not only crop loss but debt default and distress asset sales in farming communities that borrow against the harvest. A reliable 72-hour warning is therefore as much a financial intervention as a hydrological product.

Bangladesh's Flood Forecasting and Warning Centre (FFWC) issues effective bulletins for the Ganges--Brahmaputra--Meghna trunk rivers, but the haors sit largely outside this coverage. Inundation accumulates diffusely from multiple concurrent sources, including transboundary rivers that Bangladesh does not directly monitor, and fewer than twelve working gauges cover the region; during the 2017 flash flood, three of the nearest upstream gauges were inoperative. Four things break when standard flood prediction tools meet this basin:
Gauge dependency comes first: models that need dense gauge input fail exactly when conditions are worst. Seasonal confounding is subtler; raw air temperature correlates with flood labels at $r=0.570$ and with calendar month at $r=0.400$ in our dataset, so models trained on it learn the monsoon calendar, not the flood. Cloud-penetrating Sentinel-1 radar, free since 2014, is used almost only for mapping floods after they happen rather than as a pre-event predictive input. And research models rarely reach farmers as actionable, Bengali-language alerts.

This paper makes four contributions:
\begin{itemize}\setlength\itemsep{0pt}
\item a deseasonalization pipeline that removes calendar-driven bias from the feature space;
\item SAR backscatter and a modeled upstream Barak discharge proxy (${\sim}36$~h lead) as predictive inputs;
\item a 90.9\%-accurate RF+XGB ensemble, validated by LOOCV on 77 real-SAR events through a seeded, re-runnable evaluation, cross-checked against independent FFWC gauge records and a live prospective run;
\item an end-to-end Bengali alert pipeline with scenario-based boro rice damage estimation.
\end{itemize}

\section{Related Work}
Prior research relevant to haor flood warning falls into four streams. We review each and then state the gap this work fills.

\subsection{Remote sensing and SAR-based flood monitoring}
Radar satellites see through the clouds that blind optical sensors during storm season. Operational Sentinel-1 flood mapping over Bangladesh reached 89\% accuracy in \cite{uddin2019}, split-based SAR thresholding was formalized in \cite{chini2017}. Strong work, all of it. Yet each of these systems maps water that is already on the ground; none uses the SAR-derived pre-event surface state as an input for predicting the \emph{next} flood.

\subsection{Machine learning for flood prediction}
Machine learning and deep learning were applied to historical Bangladeshi climate records in \cite{rajab2023}, though that pipeline still depends on station inputs. ANN susceptibility maps for Sylhet Division were built in \cite{rudra2023}, and national-scale ensemble and stacking studies \cite{rahman2019,rahman2021,islam2021,hasan2023,chowdhury2024} report accuracies of 80--90\%. Most recently, \cite{chowdhury2025} evaluated ANN, RNN, a Random Forest and Gradient Boosting hybrid, and CatBoost for flash flood susceptibility across eight northeastern districts using Sentinel-1 SAR in Google Earth Engine, noting that satellite studies in this region often misread permanent haor water as flood water. That work shares our region and sensor stack yet remains spatial susceptibility mapping, not temporal forecasting. These accuracies deserve caution. None of these studies tests for calendar-season bias, and recent work continues to feed raw temperature into models without any deconfounding step \cite{toufique2024}, even though deseasonalization has long been standard in classical hydrology \cite{montanari2005}. Deep learning brings its own trap at this data scale: an LSTM trained on an earlier cohort of our inventory scored 100\% walk-forward accuracy (AUC 1.000), a memorization signature consistent with known data-hungry LSTM behavior \cite{kratzert2018}, so we exclude it.

\subsection{Hydrological setting of the haor region}
The roughly 36-hour travel time of the Barak River from Silchar, Assam, to the haor was measured in \cite{dewan2015}. This physical lead time is well documented and widely cited. No prior machine-learning pipeline for the region has built it in as a predictive signal.

\subsection{Early warning systems and last-mile delivery}
Evidence in \cite{perera2020} shows that last-mile message delivery and community trust decide whether a warning system saves crops, more than model accuracy does. Bangladesh's FFWC bulletins reach district offices for the major rivers, but no haor-specific forecast reaches farming communities directly.

\subsection{Research gap and positioning}
\begin{table}[!b]
\caption{Capability comparison with representative prior studies}
\label{tab:gap}
\centering
\begin{tabular}{lcccc}
\toprule
Study & \begin{tabular}{@{}c@{}}SAR as\\predictor\end{tabular} & \begin{tabular}{@{}c@{}}Upstream\\proxy\end{tabular} & \begin{tabular}{@{}c@{}}Deconf.\end{tabular} & \begin{tabular}{@{}c@{}}Farmer\\alerts\end{tabular}\\
\midrule
Uddin et al.\ 2019 \cite{uddin2019} & $\times$ & $\times$ & $\times$ & $\times$\\
Rajab et al.\ 2023 \cite{rajab2023} & $\times$ & $\times$ & $\times$ & $\times$\\
Rudra \& Sarkar 2023 \cite{rudra2023} & $\times$ & $\times$ & $\times$ & $\times$\\
Chowdhury et al.\ 2025 \cite{chowdhury2025} & $\times$ & $\times$ & $\times$ & $\times$\\
\textbf{This work} & $\checkmark$ & $\checkmark$ & $\checkmark$ & $\checkmark$\\
\bottomrule
\end{tabular}
\end{table}
Putting the four streams together, three gaps remain open (Table~\ref{tab:gap}). No Bangladesh haor study uses SAR backscatter as a pre-event predictive feature in a temporal forecasting model. No prior flood ML study for the region tests for or corrects calendar-season bias, which means reported accuracies are likely inflated. And no published system connects a haor flood model to a Bengali-language alert pipeline with crop damage estimates. This paper addresses all three within one verified, reproducible framework.

\section{Methodology}
\subsection{Study Area}
The Sunamganj Haor (${\sim}24.8$--$25.2^{\circ}$N, $91.2$--$91.6^{\circ}$E) is fed by the Surma, Kushiyara, and Baulai rivers and, from the northeast, the transboundary Barak River monitored at Silchar, Assam, with a ${\sim}36$-hour downstream travel time \cite{dewan2015}. Total relief across the basin is under 3~m; Topographic Wetness Index values of 14--20 mark a closed depression with no real outlet. When Barak discharge at Silchar exceeds roughly 6{,}000~m$^3$/s, downstream haor flooding follows within 36--48~h \cite{dewan2015}.

\subsection{Dataset and Processing Pipeline}
\begin{figure}[t]
\centering
\includegraphics[width=\columnwidth]{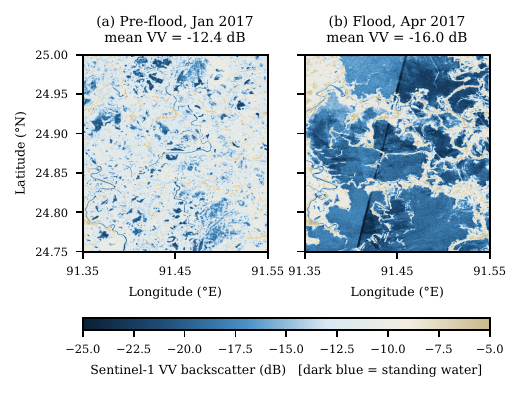}
\caption{Real Sentinel-1 VV backscatter over the Sunamganj Haor (median composites, focal-mean speckle smoothing): (a) dry-season reference, January 2017; (b) the April 2017 pre-monsoon flash flood. Mean VV drops from $-12.4$ to $-16.0$~dB as smooth standing water reflects the radar signal away from the sensor; dark tones indicate inundation.}
\label{fig:sar}
\end{figure}
The event inventory spans 2009--2024: 131 events (60 flood, 71 dry). The 77 events from 2014 onward carry real Sentinel-1 acquisitions and form the \emph{primary validation set}; all headline metrics refer to them. The 54 pre-2014 events predate the Sentinel-1 mission, so they carry physics-calibrated proxy SAR values (flooded: $-18$ to $-24$~dB; dry: $-9$ to $-14$~dB, drawn from post-2014 observed distributions). These proxies exist for one reason: they extend the training record to include the 2009--2013 flood seasons that real SAR cannot cover. Because they are constructed from post-2014 statistics, using them for evaluation would be circular, so every headline metric in this paper is computed on the 77 real-SAR events only. Labels derive from FFWC annual reports, MODIS-era inundation records \cite{islam2010}, and World Bank GRADE assessments.

Sentinel-1 GRD scenes (orbit corrections applied upstream in the Copernicus GRD product) are processed in Google Earth Engine \cite{gorelick2017} with focal-mean speckle smoothing (50~m circular kernel, a Lee-filter approximation) and conversion to dB; Fig.~\ref{fig:sar} shows the resulting backscatter contrast between dry and flooded states. Radiometric terrain correction and incidence-angle normalization are not applied; the basin's sub-3-m relief limits terrain-induced distortion, but this remains a stated limitation (Section~VI). Spatial averaging over a haor box and an upstream Silchar box (36-h lag) yields VV, VH, VV/VH ratio, and upstream$_{vv}$. Ancillary inputs include CHIRPS 7-day rainfall \cite{funk2015}, ERA5-Land soil moisture and wind \cite{munoz2021}, Open-Meteo 12-h/72-h forecast rainfall \cite{openmeteo2024}, Sentinel-2 NDWI \cite{esa2024s2} (zero-filled for the 69\% of events with $>$70\% cloud cover), and GloFAS-modeled Surma/Barak discharge \cite{glofas2024}. These discharge inputs are modeled reanalysis signals, not observed transboundary gauge readings. Event selection follows three rules: each event is a calendar date labeled flood or dry from the independent sources above, dates whose nearest Sentinel-1 acquisition exceeded 9 days were excluded so that stale imagery cannot pose as current surface state, and all feature standardization is fitted strictly within each training fold so no test information leaks into training. Processing runs in a fixed order: acquisition, speckle smoothing, dB conversion, spatial averaging, deseasonalization, then modeling.

\begin{figure}[!t]
\centering
\includegraphics[width=\columnwidth]{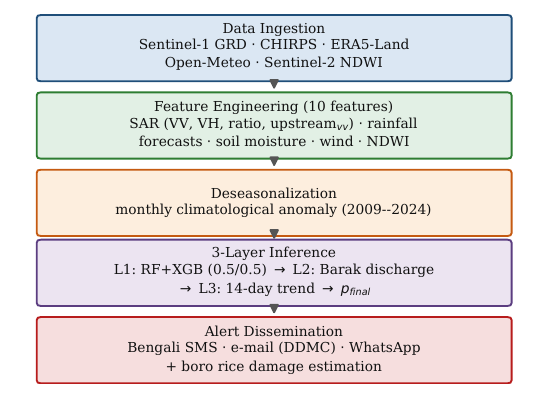}
\caption{HaorFloodAlert system architecture.}
\label{fig:arch}
\end{figure}

Fig.~\ref{fig:arch} summarizes the full pipeline: ingestion, feature engineering, deseasonalization, three-layer inference, and alert dissemination.

\subsection{Feature Set}
After dropping zero-variance variables (slope and TWI are constant in flat terrain; temp\_anomaly becomes zero-variance after deconfounding), ten features remain: 7-day CHIRPS rainfall, soil moisture, 72-h and 12-h forecast rainfall, VV, VV/VH ratio, wind speed, VH, upstream$_{vv}$, and NDWI.

\subsection{Deseasonalization}
\begin{figure}[!b]
\centering
\includegraphics[width=0.92\columnwidth]{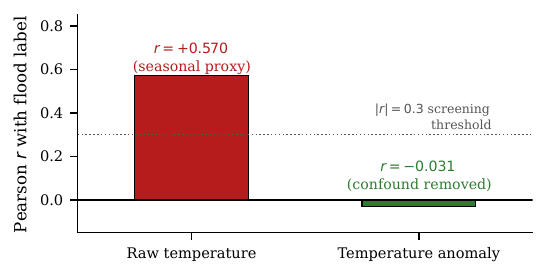}
\caption{Seasonal confound elimination for the temperature feature. Correlations computed live on the full 131-event inventory.}
\label{fig:deconf}
\end{figure}

Raw 2-m temperature correlates with flood labels at $r=0.5704$ ($p=1.15\times10^{-12}$) and with calendar month at $r=0.400$ ($p=2.21\times10^{-6}$). We substitute a monthly climatological anomaly,
\begin{equation}
\text{temp\_anomaly} = T_{\text{obs}} - T_{\text{clim}}[\text{month}],
\end{equation}
where $T_{\text{clim}}$ is the ERA5-Land 2009--2024 monthly mean. After this transform the label correlation falls to $r=-0.0305$ ($p=0.729$) (Fig.~\ref{fig:deconf}); the transformed feature then carries effectively no variance in this inventory and is excluded, so the deployed model cannot exploit the seasonal shortcut at all. We frame this as bias removal rather than an accuracy optimization, and propose a simple community standard: report the Pearson correlation of every candidate feature with calendar month; features with $|r|>0.3$ should be replaced by climatological anomalies or accuracy should be reported both raw and corrected.

\begin{figure*}[!t]
\centering
\includegraphics[width=\textwidth]{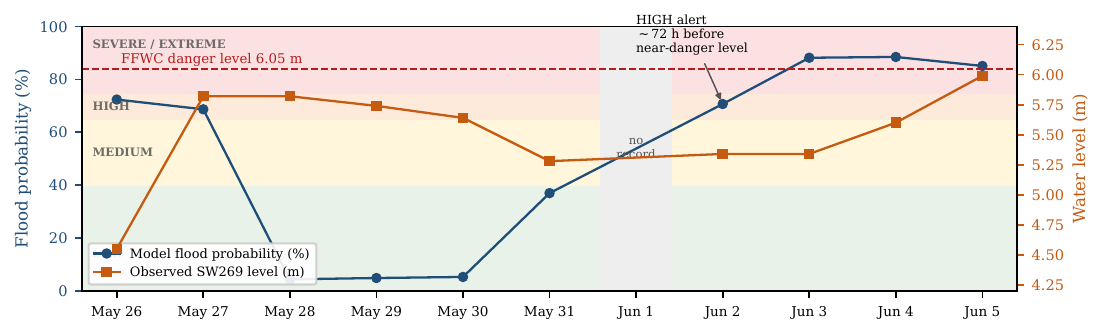}
\caption{Ten-day prospective validation (May 26--June 5, 2026): model flood probability versus independently observed FFWC SW269 water level (one representative reading per day; June 1 was not recorded). Shaded bands show operational risk classes.}
\label{fig:prosp}
\end{figure*}
\subsection{Ensemble Architecture and Training}
\textbf{Layer 1 (base, evaluated):} Random Forest \cite{breiman2001} and XGBoost \cite{chen2016} probabilities combine as an unweighted average,
\begin{equation}
p_{\text{base}} = 0.5\,p_{\text{RF}} + 0.5\,p_{\text{XGB}},
\end{equation}
with the hyperparameters of Table~\ref{tab:hyper}; all reported metrics in Section~V are computed from $p_{\text{base}}$ at decision threshold 0.5.
\textbf{Layers 2--3 (operational only):} in deployment, a discharge adjustment ($+0.10$ to $+0.15$ when GloFAS Barak discharge exceeds 6{,}000--7{,}500~m$^3$/s) and a 14-day OLS trend adjustment ($+0.05$ to $+0.15$ when $R^2\geq0.60$), capped jointly at $+0.30$ with a 0.95 probability ceiling, raise alert sensitivity during upstream buildups. These layers are heuristic post-processing for the live alerting system and are \emph{not} part of the evaluated classifier. The final probability maps to five risk classes (LOW $<$ 0.40; MEDIUM 0.40--0.65; HIGH 0.65--0.75; SEVERE 0.75--0.85; EXTREME $>$ 0.85).

Gaussian data augmentation ($8\times$, $\sigma=0.04$, applied only within each fold's training split; the held-out event is always raw) lifts LOOCV accuracy from 85.7\% to 90.9\% and F1 from 82.0\% to 89.2\% (AUC 0.922 $\rightarrow$ 0.939). The entire evaluation is seeded (seed 42) and re-runnable from a single script that writes all reported numbers, and per-event LOOCV predictions, to versioned result files released with the code.

\begin{table}[t]
\caption{Model configuration (verified against saved models)}
\label{tab:hyper}
\centering
\begin{tabular}{ll}
\toprule
Parameter & Value\\
\midrule
RF: trees / max depth / min split & 500 / 5 / 2\\
RF: class weighting & balanced\\
XGB: estimators / max depth / learning rate & 500 / 4 / 0.05\\
XGB: class imbalance & per-fold scale\_pos\_weight\\
Ensemble weights (RF / XGB) & 0.5 / 0.5\\
Decision threshold & 0.5\\
Augmentation (train folds only) & $8\times$, $\sigma=0.04$\\
Random seed & 42\\
\bottomrule
\end{tabular}
\end{table}

\subsection{Alert and Crop Damage Modules}
MEDIUM risk sends an advisory Bengali SMS (BulkSMSBD, 160 characters); HIGH adds e-mail reports to district officials; SEVERE/EXTREME activates WhatsApp-ready Bengali templates and DDMC escalation. A scenario-based damage module estimates boro rice loss as $D = D_{\text{stage}} \times F_{\text{depth}} \times F_{\text{duration}}$, intersecting Otsu-derived \cite{otsu1979} flood extent with crop-area assumptions and BRRI-informed stage-loss fractions; outputs carry $\pm$25--40\% uncertainty and are planning figures, not audited loss totals.

\section{Results}

\subsection{Validation Protocol}
Primary: LOOCV on the 77 real-SAR events (each event predicted by a model trained on the remaining 76). Secondary: 5-fold stratified CV on both the full 131-event inventory and the 77-event subset. Tertiary: stratified 60/40 holdout over five random seeds. Independent: FFWC gauge cross-checking and a 10-day prospective run. All numbers below are produced by a single seeded evaluation script and are traceable to its released output files.

\subsection{Primary LOOCV Performance}
\begin{figure}[!t]
\centering
\includegraphics[width=\columnwidth]{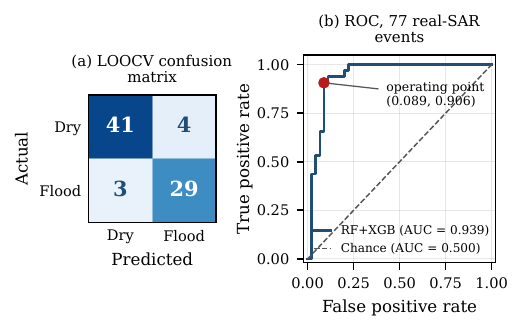}
\caption{(a) LOOCV confusion matrix (TN=41, FP=4, FN=3, TP=29) and (b) ROC curve computed from the released per-event predictions.}
\label{fig:cmroc}
\end{figure}

All reported metrics correspond to the RF+XGB base ensemble (Layer~1); the discharge and trend layers adjust operational alert probabilities only and are not part of the evaluated classifier. Table~\ref{tab:loocv} and Fig.~\ref{fig:cmroc} report the primary result. Out of 77 events the ensemble classified 70 correctly (90.9\%), missing 3 floods and raising 4 false alarms, with AUC 0.939. The ROC curve in Fig.~\ref{fig:cmroc}(b) is plotted directly from the released per-event LOOCV predictions.

\begin{table}[t]
\caption{Primary LOOCV performance, 77 real-SAR events (threshold 0.5)}
\label{tab:loocv}
\centering
\begin{tabular}{lcc}
\toprule
Metric & Value & Derivation\\
\midrule
Accuracy & 90.9\% & 70/77 correct\\
Precision & 87.9\% & 29/33 predicted floods correct\\
Recall & 90.6\% & 29/32 actual floods detected\\
F1-score & 89.2\% & harmonic mean\\
Specificity & 91.1\% & 41/45 dry events correct\\
AUC-ROC & 0.939 & from per-event predictions\\
\bottomrule
\end{tabular}
\end{table}

\subsection{Baselines, Features, and Ablation}
\begin{figure}[!t]
\centering
\includegraphics[width=\columnwidth]{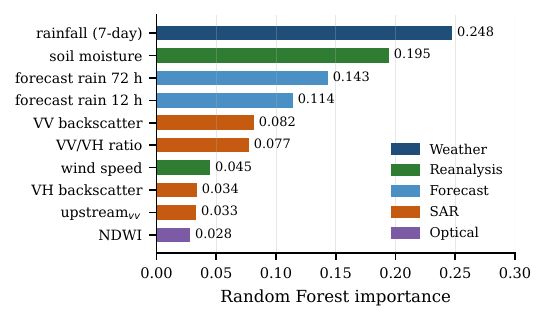}
\caption{Random Forest feature importance, read from the deployed model (10 active features).}
\label{fig:imp}
\end{figure}

\begin{figure}[t]
\centering
\includegraphics[width=\columnwidth]{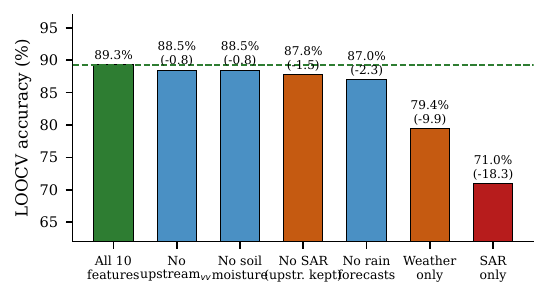}
\caption{Feature-group ablation, LOOCV on the full 131-event inventory. Deltas relative to the 89.3\% all-feature baseline.}
\label{fig:abl}
\end{figure}

Table~\ref{tab:base} shows the ensemble outperforming a majority classifier, a rainfall-threshold rule, and logistic regression evaluated under the identical LOOCV protocol. Feature importances (Fig.~\ref{fig:imp}), read directly from the deployed Random Forest, are physically interpretable: the rainfall group dominates (50.5\% combined), consistent with a basin that floods by regional accumulation, while the SAR group contributes 22.7\%. The upstream$_{vv}$ importance (0.033) understates its value, because its ${\sim}36$-h lead time cannot be expressed in an importance score. In the ablation over the full 131-event inventory (Fig.~\ref{fig:abl}), removing the three haor SAR channels costs 1.5~pp even with upstream$_{vv}$ retained, removing rain forecasts costs 2.3~pp, and single-source configurations collapse (SAR-only 71.0\%, weather-only 79.4\%): no single data stream suffices. Each feature group is justified by what breaks without it: drop upstream$_{vv}$ and accuracy falls only 0.8~pp but the system loses its only source of 36-hour lead time from a river Bangladesh cannot gauge; drop the rain forecasts and 2.3~pp goes along with the model's forward view of the sky; drop soil moisture and the antecedent wetness that primes backwater flooding vanishes. Accuracy understates these roles, which is why the ablation, not the importance ranking, carries the justification.

\begin{table}[t]
\caption{Baseline comparison, 77-event real-SAR LOOCV}
\label{tab:base}
\centering
\begin{tabular}{lcc}
\toprule
Model & Accuracy & AUC\\
\midrule
Majority classifier & 58.4\% & --\\
Rainfall threshold rule & 75.3\% & 0.799$^{\dagger}$\\
Logistic regression & 80.5\% & 0.912\\
\textbf{RF+XGB ensemble} & \textbf{90.9\%} & \textbf{0.939}\\
\bottomrule
\multicolumn{3}{l}{\footnotesize $^{\dagger}$Using raw 7-day rainfall as the ranking score.}
\end{tabular}
\end{table}

\subsection{Stability, Overfitting Control, and Generalization}
Overfitting is controlled by design, not assertion: LOOCV holds out each event completely, augmentation touches only training folds, standardization is refit inside every fold, deconfounding removes the feature that let models memorize the calendar, and the LSTM was excluded because its perfect walk-forward score was memorization. Generalization is then measured, not assumed. Five-fold stratified CV gives $88.5\% \pm 4.3\%$ accuracy (AUC $0.946 \pm 0.028$) on the full 131-event inventory and $88.2\% \pm 6.5\%$ on the 77-event real-SAR subset. The stratified 60/40 holdout over five seeds averages $81.3\% \pm 6.6\%$ accuracy with AUC $0.918 \pm 0.049$; this drop relative to LOOCV is expected: only about 46 events remain for training in each split, so the gap measures the cost of small data, not hidden memorization. A model that had merely memorized its training set would collapse far harder under this test, and the physically sensible feature ranking in Fig.~\ref{fig:imp} points the same way. Fold-level variance of this magnitude is why per-prediction ensemble-variance flags route uncertain cases to manual DDMC review in deployment.

\subsection{Independent FFWC Cross-Checking}
Joining the 22{,}473 three-hourly readings (2014--2026) of FFWC station SW269 \cite{ffwc2026} to the labeled inventory by calendar date matches 97 of 131 events. Flood-labeled dates average 5.77~m water level ($n=43$) versus 2.76~m for dry dates ($n=54$); the two groups separate cleanly ($t=6.77$, $p=1.1\times10^{-9}$; point-biserial $r=0.562$). A naive single-threshold gauge rule (best in-sample threshold 2.65~m) reaches 76.3\% on these matched dates, an in-sample descriptive figure, not a cross-validated claim, indicating that the ensemble's cross-validated 90.9\% reflects predictive skill beyond what a single water-level reading provides,.

In a prospective run (May 26--June 5, 2026; Fig.~\ref{fig:prosp}) against live SW269 readings \cite{ffwc2026}, the system flagged HIGH risk on May 26 while the gauge sat 150~cm below danger, one day before a sharp rise to within 23~cm of danger. The main mismatch occurred on May 28: probability dropped to 4.4\% while water remained high, because declining forecast rainfall drove the model toward recession faster than the physical gauge. That behavior is useful for early all-clear signals but risky if alerts relax too soon. In the second buildup, the June 2 HIGH alert preceded the June 5 near-danger level (6~cm below the 6.05-m mark) by roughly 72 hours: 8 clearly consistent days plus one borderline day out of 10.

\section{Limitations}
The 77-event real-SAR inventory is small relative to the ${\sim}200$ events suggested for stable Random Forests \cite{breiman2001}; pre-2014 proxies carry circularity risk, so we keep them out of the primary metrics. NDWI zero-filling injects a systematic signal during cloud-covered high-risk periods; and GloFAS reanalysis thresholds may not transfer directly to real-time observed discharge. The GEE SAR chain applies focal-mean speckle smoothing but no radiometric terrain correction or incidence-angle normalization; the basin's near-zero relief mitigates, but does not eliminate, this gap. The holdout results quantify the accuracy cost of the small-sample regime. The prospective run is a short pilot, not a seasonal campaign. Finally, alert errors carry asymmetric social costs: a false SEVERE alert can trigger costly premature harvesting, while the May~28 early-recession mismatch illustrates the inverse risk of premature all-clears; deployment therefore requires conservative de-escalation rules and human review of borderline cases, and alert contact lists must be handled under appropriate data-privacy safeguards.

\section{Conclusion}
HaorFloodAlert shows that deseasonalized, SAR-informed ensemble learning can produce credible 72-hour flash flood forecasts in a wetland with almost no gauges, and can carry a forecast from satellite acquisition to a Bengali-language farmer alert in about two hours of processing, leaving an effective warning window of roughly 34 hours with the Barak proxy. Every reported number is regenerated by a single seeded script released with the code, including per-event predictions. Future work targets expansion to 200+ verified real-SAR events, physics-informed augmentation, a 12-month single-upazila field trial, real-time Barak discharge through a CWC data-sharing arrangement, and climate-adaptive rolling retraining of the climatological baseline as pre-monsoon precipitation patterns shift under climate change. The measure of this work is not a table of metrics but whether a farmer in Sunamganj, holding a basic phone, receives a Bengali warning early enough to act.

\section*{Acknowledgment}
The authors thank BWDB and FFWC for gauge data (station SW269), ESA for Sentinel-1 imagery, and NASA for precipitation products. Code, the evaluation pipeline, and per-event predictions are available at \url{https://github.com/shkoli/HaorFloodAlert}.


\begin{thebibliography}{00}
\bibitem{kamal2018} A.~S.~M.~M. Kamal \emph{et al.}, ``Resilience to flash floods in wetland communities of northeastern Bangladesh,'' \emph{Int. J. Disaster Risk Reduct.}, vol.~31, pp. 478--488, 2018.
\bibitem{uddin2019} K. Uddin, M.~A. Matin, and F.~J. Meyer, ``Operational flood mapping using multi-temporal Sentinel-1 SAR images: A case study from Bangladesh,'' \emph{Remote Sens.}, vol.~11, no.~13, p. 1581, 2019.
\bibitem{chini2017} M. Chini, R. Hostache, L. Giustarini, and P. Matgen, ``A hierarchical split-based approach for parametric thresholding of SAR images: Flood inundation as a test case,'' \emph{IEEE Trans. Geosci. Remote Sens.}, vol.~55, no.~12, pp. 6975--6988, 2017.
\bibitem{rajab2023} A. Rajab \emph{et al.}, ``Flood forecasting by using machine learning: A study leveraging historic climatic records of Bangladesh,'' \emph{Water}, vol.~15, no.~22, p. 3970, 2023.
\bibitem{rudra2023} R.~R. Rudra and S.~K. Sarkar, ``Artificial neural network for flood susceptibility mapping in Bangladesh,'' \emph{Heliyon}, vol.~9, no.~6, e16459, 2023.
\bibitem{rahman2019} M. Rahman \emph{et al.}, ``Flood susceptibility assessment in Bangladesh using machine learning and multi-criteria decision analysis,'' \emph{Earth Syst. Environ.}, vol.~3, pp. 585--601, 2019.
\bibitem{rahman2021} M. Rahman \emph{et al.}, ``Application of stacking hybrid machine learning algorithms in delineating multi-type flooding in Bangladesh,'' \emph{J. Environ. Manage.}, vol.~295, 113086, 2021.
\bibitem{islam2021} A.~R.~M.~T. Islam \emph{et al.}, ``Flood susceptibility modelling using advanced ensemble machine learning models,'' \emph{Geosci. Front.}, vol.~12, no.~3, 101075, 2021.
\bibitem{hasan2023} M.~H. Hasan, A. Ahmed, K.~M. Nafee, and M.~A. Hossen, ``Use of machine learning algorithms to assess flood susceptibility in the coastal area of Bangladesh,'' \emph{Ocean Coast. Manage.}, vol.~236, 106503, 2023.
\bibitem{chowdhury2024} M.~S. Chowdhury, ``Flash flood susceptibility mapping of north-east depression of Bangladesh using different GIS based bivariate statistical models,'' \emph{Watershed Ecol. Environ.}, vol.~6, pp. 26--40, 2024.
\bibitem{chowdhury2025} M.~E. Chowdhury, A.~K.~M.~S. Islam, R.~U. Zaman, and S. Khadem, ``A machine learning-based approach for flash flood susceptibility mapping considering rainfall extremes in the northeast region of Bangladesh,'' \emph{Adv. Space Res.}, vol.~75, 2025, doi: 10.1016/j.asr.2024.10.047.
\bibitem{toufique2024} S.~M. Toufique \emph{et al.}, ``Implementing machine learning techniques to forecast floods in Bangladesh,'' in \emph{Proc. 4th Int. Conf. Electr., Comput. Energy Technol. (ICECET)}, 2024, pp. 1--6.
\bibitem{montanari2005} A. Montanari, ``Deseasonalisation of hydrological time series through the normal quantile transform,'' \emph{J. Hydrol.}, vol.~313, no.~3--4, pp. 274--282, 2005.
\bibitem{kratzert2018} F. Kratzert, D. Klotz, C. Brenner, K. Schulz, and M. Herrnegger, ``Rainfall--runoff modelling using long short-term memory (LSTM) networks,'' \emph{Hydrol. Earth Syst. Sci.}, vol.~22, pp. 6005--6022, 2018.
\bibitem{dewan2015} A.~M. Dewan, M.~M. Islam, T. Kumamoto, and M. Nishigaki, ``Barak--Surma--Meghna river system: Flood hazard assessment using geospatial techniques,'' \emph{Geomat. Nat. Hazards Risk}, vol.~6, suppl.~1, pp. 1--15, 2015.
\bibitem{perera2020} D. Perera, J. Agnihotri, O. Seidou, and R. Djalante, ``Identifying societal challenges in flood early warning systems,'' \emph{Int. J. Disaster Risk Reduct.}, vol.~51, 101794, 2020.
\bibitem{islam2010} A.~S.~M. Islam, S.~K. Bala, and M.~A. Haque, ``Flood inundation map of Bangladesh using MODIS time-series images,'' \emph{J. Flood Risk Manage.}, vol.~3, no.~3, pp. 210--222, 2010.
\bibitem{gorelick2017} N. Gorelick \emph{et al.}, ``Google Earth Engine: Planetary-scale geospatial analysis for everyone,'' \emph{Remote Sens. Environ.}, vol.~202, pp. 18--27, 2017.
\bibitem{funk2015} C. Funk \emph{et al.}, ``The climate hazards infrared precipitation with stations, a new environmental record for monitoring extremes,'' \emph{Sci. Data}, vol.~2, 150066, 2015.
\bibitem{munoz2021} J. Mu\~noz-Sabater \emph{et al.}, ``ERA5-Land: A state-of-the-art global reanalysis dataset for land applications,'' \emph{Earth Syst. Sci. Data}, vol.~13, no.~9, pp. 4349--4383, 2021.
\bibitem{openmeteo2024} Open-Meteo, ``Weather and flood API documentation,'' 2024. [Online]. Available: \url{https://open-meteo.com/en/docs}
\bibitem{esa2024s2} European Space Agency, ``Sentinel-2 MSI data,'' Copernicus Data Space Ecosystem, 2024.
\bibitem{glofas2024} Copernicus Emergency Management Service, ``River discharge and related historical data from the Global Flood Awareness System (GloFAS),'' Copernicus Early Warning Data Store, 2024.
\bibitem{breiman2001} L. Breiman, ``Random forests,'' \emph{Mach. Learn.}, vol.~45, no.~1, pp. 5--32, 2001.
\bibitem{chen2016} T. Chen and C. Guestrin, ``XGBoost: A scalable tree boosting system,'' in \emph{Proc. 22nd ACM SIGKDD Int. Conf. Knowl. Discov. Data Min.}, 2016, pp. 785--794.
\bibitem{otsu1979} N. Otsu, ``A threshold selection method from gray-level histograms,'' \emph{IEEE Trans. Syst., Man, Cybern.}, vol.~9, no.~1, pp. 62--66, 1979.
\bibitem{ffwc2026} Bangladesh Water Development Board, Flood Forecasting and Warning Centre, ``Water level data for Sunamganj station SW269,'' 2026. [Online]. Available: \url{https://old.ffwc.gov.bd}
\end{thebibliography}
\end{document}